# The threshold EM algorithm for parameter learning in bayesian network with incomplete data

Fradj Ben Lamine
Department of computer science
Faculty of science of Monastir
Monastir – Tunisia
Benlamine.fradj@gmail.com

Karim Kalti
Department of computer science
Faculty of science of Monastir
Monastir – Tunisia
karim.kalti@gmail.com

Mohamed Ali Mahjoub
Preparatory Institute of Engineer of Monastir
Monastir – Tunisia
medali.mahjoub@ipeim.rnu.tn

*Abstract*—Bayesian networks (BN) are used in a big range of applications but they have one issue concerning parameter learning. In real application, training data are always incomplete or some nodes are hidden. To deal with this problem many learning parameter algorithms are suggested foreground EM, Gibbs sampling and RBE algorithms. In order to limit the search space and escape from local maxima produced by executing EM algorithm, this paper presents a learning parameter algorithm that is a fusion of EM and RBE algorithms. This algorithm incorporates the range of a parameter into the EM algorithm. This range is calculated by the first step of RBE algorithm allowing a regularization of each parameter in bayesian network after the maximization step of the EM algorithm. The threshold EM algorithm is applied in brain tumor diagnosis and show some advantages and disadvantages over the EM algorithm.

## I. INTRODUCTION

Machine Learning is now considered among the essential tools for making decisions and solving problems that affect the uncertainty. This science allows automation of methods that helps the expert to take an effective decision in several areas. This work is functional by means of artificial intelligence that combines the concepts of learning, reasoning and problem-solving. In recent years, Bayesian networks have become important tools for modeling uncertain knowledge. They are used in various applications such as information retrieval [6,14], data fusion [5], bioinformatics [11], classification [12,13] and medical diagnostics [2].

Bayesian networks are graphical models that can apply these concepts in daily life by modeling a given problem as a causal structure as a graph indicating the independence between the different actors of the problem and using qualitative state which is in the form of conditional probability tables. The clarity of the semantics and comprehensibility by humans are the major advantages of using Bayesian networks for modeling applications. They offer the possibility of causal interpretation of models of learning.

The concepts of learning in bayesian network are devised into two types; the first one is to learn the parameters when the structure is known. The second one is to learn the structure and the parameters at the same moment. In this paper, we assume that the structure is known. The parameter learning in this case is divided into two categories. If the training data are complete this problem is resolved by statistic approach or a bayesian approach. In real application, to find complete training data is difficult for various reasons. When data are incomplete two classical approaches are usually used to determine the parameters of a bayesian network that include EM algorithm [1] and Gibbs Sampling [3].

Other methods are suggested to deal with the disadvantages of these classical approaches. The most robust is the RBE algorithm [8]. In order to regularize the learning problem, some modifications are needed to reduce the search space and help escape from local maxima.

These problems in learning parameter in bayesian network motivate us to add some modification in the existing parameter learning algorithm where the network structure is known and the data are incomplete.

## II. LEARNING BAYESIAN NETWORK PARAMETERS

A bayesian network is defined by a set of variables $\chi = \{X_1, X_2, ..., X_n\}$ that represent the actors of the problem and a set of edge that represent the conditional independence between these variables. If there is an arc from $X_i$ to $X_j$ then $X_i$ is called parent of $X_j$ and is noted by $pa(X_j)$. Each node is conditionally independent from all the other nodes given its parents. The conditional distribution of all nodes is described as:

$$P(\chi) = \prod_{i=1}^{n} P(X_i | pa(X_i)) \qquad (1)$$

Each node is described by a conditional probability table which we denote by the vector θ. The entire vector is composed by a set of parameters value $\theta_{i,j,k}$ and it's defined by:

$$\theta_{i,j,k} = P(X_i = x_k | pa(X_i) = x_j) \qquad (2)$$

Where i=1…n represents the range of all variables, k=1…$r_i$ describes all possible states taken by $X_i$ and j=1...$q_i$ ranges all possible parent configurations of node $X_i$.

The process of learning parameters in bayesian network is discussed in many papers. The goal of parameter learning is to find the most probable θ that explain the data. Let $D = \{D_1, D_2,…, D_N\}$ be a training data where $D_l = \{x_1[l], x_2[l],…,x_n[l]\}$ consists of instances of the bayesian network nodes. Parameter learning is quantified by the log-likelihood function denoted as $L_D(\theta)$. When the data are complete, we get the following equations:

$$L_D(\theta) = \log\left\{\prod_{l=1}^{N} P(x_1[l], x_2[l],…,x_n[l]:\theta)\right\} \quad (3)$$

$$L_D(\theta) = \log\left\{\prod_{i=1}^{n}\prod_{l=1}^{N} P(x_i[l] | pa(x_i[l]):\theta)\right\} \quad (4)$$

The equation (3) and (4) are not applied where the training data is incomplete.

*A. Learning parameter with complete data*

In the case where all variables are observed, the simplest method and most used is the statistical estimate. It estimats the probability of an event by the frequency of occurrence of the event in the database. This approach (called maximum likelihood (ML)) then gives us:

$$P(X_i = x_k | pa(X_i) = x_j) = \theta_{i,j,k} = \frac{N_{i,j,k}}{\sum_k N_{i,j,k}} \quad (5)$$

Where $N_{i,j,k}$ is the number of events in the database for which the variable $X_i$ is in state $x_k$ and his parents are in the configuration $x_j$.

The principle, somewhat different, the Bayesian estimation is to find parameters most likely knowing that the data were observed. Using a Dirichlet distribution as a priori parameters which are written as:

$$P(\theta) \propto \prod_{i=1}^{n}\prod_{j=1}^{qi}\prod_{k=1}^{ri} \theta_{i,j,k}^{(\alpha_{i,j,k}-1)} \quad (6)$$

where $\alpha_{i,j,k}$ are the parameters of the Dirichlet distribution associated with the prior distribution.

The approach to maximum a posteriori (MAP) gives us:

$$P(X_i = x_k | pa(X_i) = x_j) = \theta_{i,j,k} = \frac{N_{i,j,k} + \alpha_{i,j,k} - 1}{\sum_k N_{i,j,k} + \alpha_{i,j,k} - 1} \quad (7)$$

*B. Learning parameter with incomplete data*

In most applications, databases are often incomplete. Some variables are observed only partially or never. The classical approaches are EM, Gibbs sampling and RBE algorithms. These algorithms are approximate except RBE which determinate a low bound and an upper bound for each parameter in the bayesian network.

The method of parameter estimation with incomplete data and the most commonly used is based on the iterative Expectation-Maximization (EM) proposed by Dempster [1] and applied to the RB in [7].

The EM above is as follows: repeat the steps expectation and maximization until the convergence. Each iteration ensures that the likelihood function increases and eventually converges to a local maximum. By cons, when we have multiple nodes admitting a large number of missing data, the method of learning by the EM method converges quickly to a local maximum. In the first step, the algorithm starts by depending arbitrary quantities on missing data. The second steps consist of employing the expectation entries and maximizing them with respect to the unknown parameters. The results of the second step are used as arbitrary quantities in the next expectation step. The algorithm converges when the difference between successive estimates is smaller than a fixed threshold or the number of iterations is bigger than a fixed maximum iteration.

*Algorithm Expectation Maximization EM (input : DAG, data base D, E function that calculate expectation)*
*output : $\theta_{i,j,k}$*
*Begin*
    *1. t=0*
    *2. Randomly initialize the parameters*
    *3. Repeat*
    *4. Expectation*
        *use the current parameters $\theta_{i,j,k}^{(t)}$ to estimate missing parameters :*
        *$E^{(t)}(N_{i,j,k}) = \Sigma\, p^{(t)}(X_i=x_k | pa(X_i)=x_j)$*
    *For i from 1 to N*
    *5. Maximization*
        *use estimate date to apply the learning procedure*
        *(for example the maximum likelihood)*
        *$\theta_{i,j,k}^{(t+1)} = E(N_{i,j,k}) / E(N_{i,j})$*
    *6. t=t+1*
    *Until convergence ($\theta_{i,j,k}^{(t+1)} = \theta_{i,j,k}^{(t)}$)*
*End*

Algorithm : EM algorithm

The second algorithm is Gibbs sampling [3] introduced by Heckerman. Gibbs sampling is described as a general method for probabilistic inference. It can be applied in all type of graphical models whether the arcs are directed or not and whether the variables are discrete or continuous. Gibbs sampling is a special case of MCMC (Markov Chain Monte Carlo). It generates a string of samples with accepting or rejecting some interesting points. In other words, Gibbs sampling consists in completing the sample by inferring the missing data from the available information. In learning the parameters, Gibbs sampling is a method that converges slowly or has no solution if the number of hidden variables is very large.

The third algorithm is Robust Bayesian Estimator RBE [8]. It's composed of two steps Bound and Collapse [10]. The first step consists of calculating a lower bound and an upper bound for each parameter in the bayesian network. The second step uses a convex combination to determine the value of $\theta_{i,j,k}$.

RBE is considered a procedure that runs through all the data D recorded observations about the variables and then it allows to bound the conditional probability of a variable Xi. This procedure begins by identifying the virtual frequencies following:

- $n(X_i = x_k | ?)$: calculating the number of observations where the variable $X_i$ takes the value $x_k$ and the value of pa $(X_i)$ is not completely observed.
- $n(? | pa(X_i) = x_j)$ calculating the number of observations where parents pa $(X_i)$ takes the value $x_j$ and the value of $X_i$ is missing.
- $n(? | ?)$: calculating the number of observations where both values of $X_i$ and pa $(X_i)$ are unknown and the value of pa$(X_i)$ can be completed as $x_j$.

These frequencies help us to calculate the minimum and maximum number of observations that may have characteristics $X_i = x_k$ and pa $(X_i) = x_j$ in the database D:

$$n_{min} = n(?|pa(X_i)=x_j) + n(X_i=x_k|?) + n(?|?) \quad (8)$$

is the minimum number of observations with characteristics $X_i = x_k$ and pa $(X_i) = x_j$.

$$n_{max} = n(?|pa(X_i)=x_j) + \sum_{(h \neq k)} n(X_i=x_k|?) + n(?|?) \quad (9)$$

is the maximum number of observations with characteristics $X_i = x_k$ and pa $(X_i) = x_j$.

Virtual frequencies defined above can be set to zero, which is called the Dirichlet distribution with parameters $\alpha_{i,j,k}$. We define the lower bound of the interval by:

$$\min_{i,j,k} = \frac{\alpha_{i,j,k} + n(X_i = x_k | pa(X_i) = x_j)}{\alpha_{i,j} + n(pa(X_i) = x_j)) + n_{min}} \quad (10)$$

And the upper bound by:

$$\max_{i,j,k} = \frac{\alpha_{i,j,k} + n(X_i = x_k | pa(X_i) = x_j) + n_{max}}{\alpha_{i,j} + n(pa(X_i) = x_j) + n_{max})} \quad (11)$$

A detailed example mentioned in [8] shows the use of these equations in calculating conditional probabilities by determining the minimum and maximum bounds of the interval. This phase of determining $\min_{i,j,k}$ and $\max_{i,j,k}$ depends only on the frequency of observed data in the database and virtual frequencies calculated by completing the records. The major advantage of this method is the independence of the distribution of missing data without trying to infer.

To find the best parameters for this method, a second phase is necessary. It estimates the parameters using a convex combination from each distribution calculated for each given node. This convex combination can be determined either by external knowledge about the missing data, or by a dynamic estimate based on valid information in the database. A description of the execution of this phase is articulated in [10].

### III. THE THRESHOLD EM ALGORITHM FOR PARAMETER LEARNING IN BAYESIAN NETWORK WITH INCOMPLETE DATA

The set of parameter in bayesian network using EM algorithm is approximate. In addition, the use of the bound step of the RBE algorithm gives a lower bound and an upper bound for each parameter in the network which is defined by :

$$\min_{i,j,k} <= \theta_{i,j,k} <= \max_{i,j,k} \quad (12)$$

Our work consists of performing the optimization of the bayesain network parameter using the EM algorithm and verifying the bound step of the RBE algorithm.

For doing that, the threshold EM algorithm consists of verifying the constrain mentionned in equation (12) after the two steps of the EM algorithm. Let $\theta_{i,j,k}^{(t)}$ be the maximized parameter after the execution of the two steps of the EM algorithm. The threshold EM algorithm is composed by three steps. The first two steps are the same as the EM algorithm. The third step consists of the regularization of $\theta_{i,j,k}^{(t)}$ with the constraint mentionned in equation (12). The main actions used in this step consists of:

i) If $\theta_{i,j,k}^{(t)} <= \min_{i,j,k}$ then the $\theta_{i,j,k}^{(t)}$ is equal to $\min_{i,j,k}$.

ii) If $\theta_{i,j,k}^{(t)} >= \max_{i,j,k}$ then the $\theta_{i,j,k}^{(t)}$ is equal to $\max_{i,j,k}$.

iii) If $\min_{i,j,k} <= \theta_{i,j,k}^{(t)} <= \max_{i,j,k}$ then the $\theta_{i,j,k}^{(t)}$ is saved like it's.

These changes provide a disagree of the probabilities constraint defined in equation (13) :

$$\sum_k \theta_{i,j,k} = 1 \quad (13)$$

So, it's necessary to make a normalization step to verify the equation (13). This step is described by the use of the equation (14).

$$\theta_{i,j,k}^{(t+1)} = \frac{\theta_{i,j,k}^{(t)}}{\sum_{k'} \theta_{i,j,k'}^{(t)}} \quad (14)$$

These new calculating parameters are used like an input in the next step of the threshold algorithm. This principle is repeated until convergence. The stopping points are the same as the EM algorithm. The third step is used to force the solution to be between the bounds calculating by the bound step of the RBE algorithm. In the worst case, the solution is moving toward the directions of reducing the violations of the constraint mentionned in equation (12).

Now, we are ready to present the threshold EM algorithm for parameter learning in bayesian network with missing data as summarized in table 1.

*Repeat until it converges*
  *Step1: Expectation step to compute the conditional expectation of the log-likelihood function.*
  *Step2: Maximization step to find the parameter $\theta^{(t)}$ that maximize the log-likelihood.*
  *Step3: Regularization step to get the parameter into the interval calculating by the bound step of the RBE algorithm:*
    *For each variable i, parent configuration j, value k*
    *If $\theta_{i,j,k}^{(t)} <= min_{i,j,k}$ then $\theta_{i,j,k}^{(t)} = min_{i,j,k}$.*
    *If $\theta_{i,j,k}^{(t)} >= max_{i,j,k}$ then $\theta_{i,j,k}^{(t)} = max_{i,j,k}$.*
    *If $min_{i,j,k} <= \theta_{i,j,k}^{(t)} <= max_{i,j,k}$ then the $\theta_{i,j,k}^{(t)}$ is saved like it's.*
  *Strep4: Normalization step based on equation (14)*
  $\theta^{(t)} = \theta^{(t+1)}$
*Go to step1*
*Return $\theta^{(t)}$*

Algorithm 1. The threshold EM algorithm

We describe in table 2 an example of using the threshold algorithm in one iteration :

TABLE I.　　AN EXAMPLE

| Min | 0,0566 | 0.07 |
|---|---|---|
| Max | 0.5 | 0.5 |
| $\theta_{i,j,k}^{(t)}$ | 0.6206 | 0.3794 |
| Regularization | 0.5 | 0.3794 |
| Normalization | 0.5686 | 0.4314 |

We see that the new parameter calculating in one step reduces the violations of the constraint mentioned in equation (12).

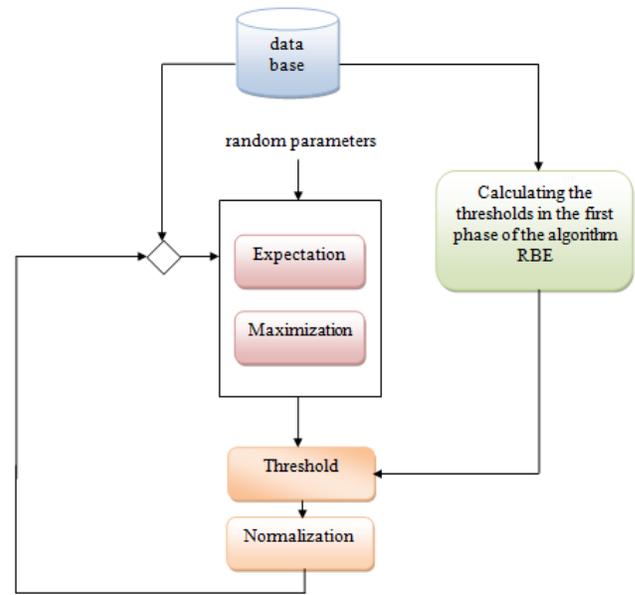

Figure 1. The threshold EM algorithm

## IV. EXPERIMENTS

During this section, we compare our algorithm to the EM algorithm. We apply this work in brain tumor diagnosis. We use the Bayesian Network Toolbooxs (BNT) by Murphy to test our algorithm. The bayesian network as shown in Fig1 is created in these experiments. Then, 72 instances are collected from a real diagnosis and we mention that not all the variables are instanced.

The dataset use to learn the bayesian network parameters is composed by 72 instances of each node tacked from a real cases collected by a specialist in brain tumor diagnosis. All these nodes are discrete and takes between two and 8 values. The percentage of the missing data in this dataset is equal to 37.16%. The majority of missing data is in the intermediate nodes of the bayesian network. The causes of the missing data are the quality of IRM images or the doctor forgot to mention all the details in this report.

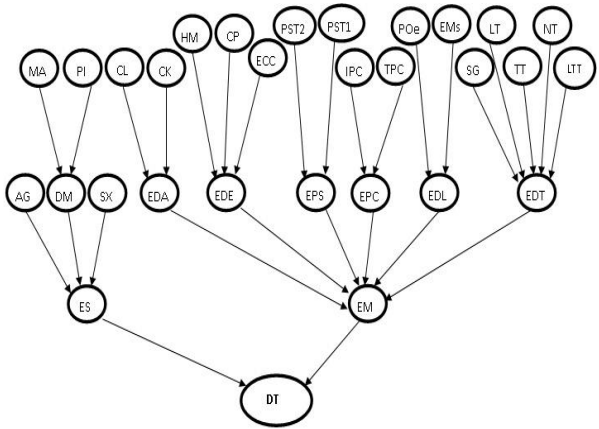

Figure 2. The Bayesian Network structure

The different meanings of names used in the figure are detailed in Table II and III.

TABLE II. TABLE TYPE STYLES

| Node name | signification |
|---|---|
| AG | Age |
| CK | Cystic component |
| CL | Calcification |
| CP | Composition |
| DM | Medical record |
| DT | Decision Tumor |
| ECC | Flooding Corpus Callosum |
| EDA | State Auxiliary Data |
| EDE | State data encephalic |
| EDL | Liquide state data |
| EDT | Tumor state data |
| EM | Radiologic state |
| Ems | Mass effect |
| EPC | State taking contrast |
| EPS | Signal taking state |

TABLE III. TABLE TYPE STYLES

| Node name | signification |
|---|---|
| ES | State clinic |
| HM | Hemorrhage |
| IPC | Importance of taking Contrast |
| LT | Tumor location |
| LTT | Tumor limit |
| MA | Diseases auxiliary |
| NT | Tumor number |
| PI | first Infection |
| Poe | Edema presence |
| PST1 | Making the Signal in T1 |
| PST2 | Making the signal in T2 |
| SG | seat |
| SX | Sex |
| TPC | Type of taking Contrast |
| TT | Tumor size |

Equation 15 allows to give the performance parameters calculated in the Bayesian network. The Log-Likelihood gives the parameters that best describe the training set. This value is updated at each iteration in the EM algorithm.

We show in this graphic that these functions are already the same. The log-likelihood of the TH_EM algorithm is lower than the log-likelihood of the EM algorithm.

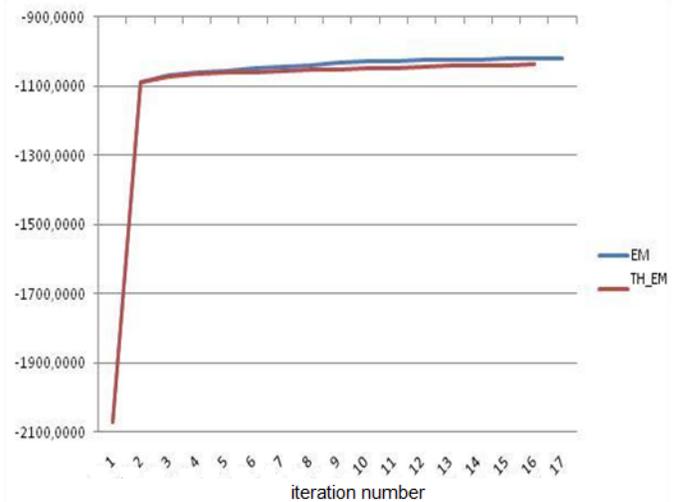

Figure 3. Comparison of the log-likelihood between TH_EM and EM algorithms

We show in Figure 3 the comparison between EM algorithm and the threshold EM algorithm (TH_EM) concerning the log-likelihood function which is defined as follows :

$$\text{LL}(D|\theta) = \log L(D|\theta) = \sum_{i=1}^{n}\sum_{j=1}^{q_i}\sum_{k=1}^{r_i} N_{i,j,k} \log \theta_{i,j,k} \quad (15)$$

where :
$n$ is the node number.
$q_i$ is node i parents configuration number.
$r_i$ the number state of node i
$N_{i,j,k}$ is the number of cases where the node i is in state k and its parents are in configuration $j$.
$\theta_{i,j,k}$ is the parameter value where node i is in state k and ists parents are in configuration $j$.

This test is applied when we fix the same starting points in the two algorithms. We see that the convergence of our algorithm is quickly than EM algorithm. This result is shown in 70% of cases when we change the starting points of the two algorithms (figure 4). In addition, we see that the probability distribution in each node is modified. Each probability is between the two bounds calculating with the first step of the RBE algorithm or error rate become smaller. One advantage of our algorithm consists of the absence of zero probability in each probability distribution.

The convex combination of the two bounds calculated in the first step of the RBE algorithm external information to get the parameters of any bayesian networks. This task becomes difficult when you have a complex structure. Our proposed method deletes the use of this information to get the conditional probability tables of our bayesian network.

## V. CONCLUSION

In real application, training data in Bayesian network are always incomplete or some nodes are hidden. Many learning parameter algorithms are suggested foreground EM, Gibbs sampling and RBE algorithms. In order to limit the search space and escape from local maxima produced by executing EM algorithm, this paper presents a learning parameter algorithm that is a fusion of EM and RBE algorithms. This algorithm incorporates the range of a parameter into the EM algorithm. The threshold EM algorithm is applied in brain tumor diagnosis and show some advantages and disadvantages over the EM algorithm

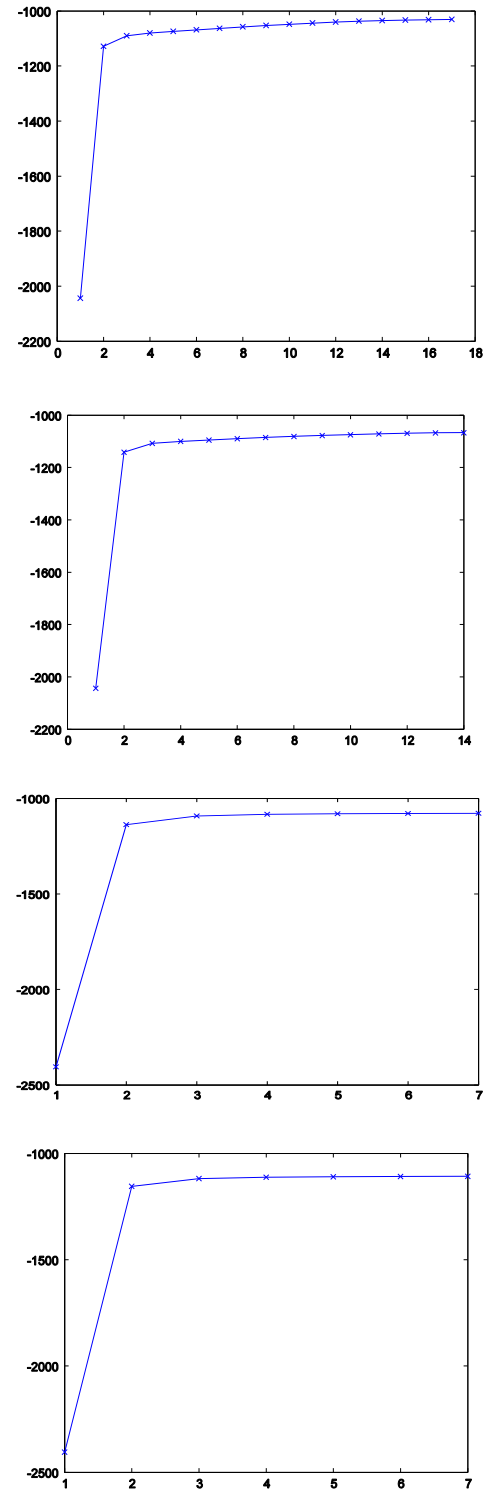

Figure 4. Log-likelihood with different starting point